\documentclass{iise}

\usepackage{multirow}
\usepackage{subfigure}
\usepackage{graphicx}
\usepackage{subcaption}

\conference{Proceedings of the IISE Annual Conference \& Expo 2025} 
\title{\titlesize Learning-Based Multi-Criteria Decision Model for Site Selection Problems}

\author{
Mahid Ahmed\thanks{Presenting Author, email: mahid.ahmed@usm.edu}, 
Ali Dogru\thanks{Email: ali.dogru@usm.edu}, 
Chaoyang Zhang\thanks{Email: chaoyang.zhang@usm.edu},
Chao Meng\thanks{Email: chao.meng@usm.edu}\\
University of Southern Mississippi\\
Hattiesburg, MS, USA
}

\authorlist{Ahmed, Dogru, Zhang ,and Meng }

\abstractID{6327}

\begin{document}
\maketitle 

\begin{abstract}
{\small Strategically locating sawmills is critical for the efficiency, profitability, and sustainability of timber supply chains, yet it involves a series of complex decision-making affected by various factors, such as proximity to resources and markets, proximity to roads and rail lines, distance from the urban area, slope, labor market, and existing sawmill data. Although conventional Multi-Criteria Decision-Making (MCDM) approaches utilize these factors while locating facilities, they are susceptible to bias since they rely heavily on expert opinions to determine the relative factor weights. Machine learning (ML) models provide an objective, data-driven alternative for site selection that derives these weights directly from the patterns in large datasets without requiring subjective weighting. Additionally, ML models autonomously identify critical features, eliminating the need for subjective feature selection. In this study, we propose integrated ML and MCDM methods and showcase the utility of this integrated model to improve sawmill location decisions via a case study in Mississippi. This integrated model is flexible and applicable to site selection problems across various industries. }
\end{abstract}

\section*{Keywords}
Multi-Criteria Decision-Making. Machine Learning, Site Location

\section{Introduction}
Sawmill location decisions are vital to the success of timber supply chains but require complex, multi-criteria decision-making (MCDM). Factors such as proximity to resources, markets, transportation infrastructure, labor availability, timber availability, existing sawmill locations, and land characteristics influence these decisions. Traditional MCDM methods often rely on expert-driven weighting of these factors, which can introduce bias. In contrast, machine learning (ML) models offer a data-driven alternative, autonomously determining factor importance from large datasets and enhancing objectivity.

This study proposes a learning-based multi-criteria decision-making (LB-MCDM) framework to address the sawmill site selection problem. We demonstrate the utility of this integrated model through a case study in Mississippi. A Random Forest classifier was used to assess six key location factors, generating a suitability map based on objective factor weights. The results highlight the model’s effectiveness in improving site selection decisions, with applications across various industries. 

This abstract is structured as follows: Section 2 reviews related literature. Section 3 formally defines the sawmill site selection problem. Section 4 outlines the methodology, detailing the use of raster data and the integration of machine learning and MCDM approaches. Section 5 presents the preliminary results. Section 6 concludes with key findings, discusses implications, and outlines the next steps.

\section{Related Literature}
Multi-Criteria Decision Making (MCDM) techniques are widely used for solving business location problems by incorporating multiple factors to support informed decision-making. These methods have applications across various fields, including healthcare, urban and regional planning, logistics, and environmental management\citep{aruldoss13}. Commonly used MCDM techniques include the analytic hierarchy process (AHP), weighted sum model (WSM), hybrid models such as fuzzy-AHP, geographic information system (GIS)-based MCDM, and data-driven approaches like machine learning models.  
\begin{figure}[!ht]
	\centering
	\includegraphics[width=\textwidth]{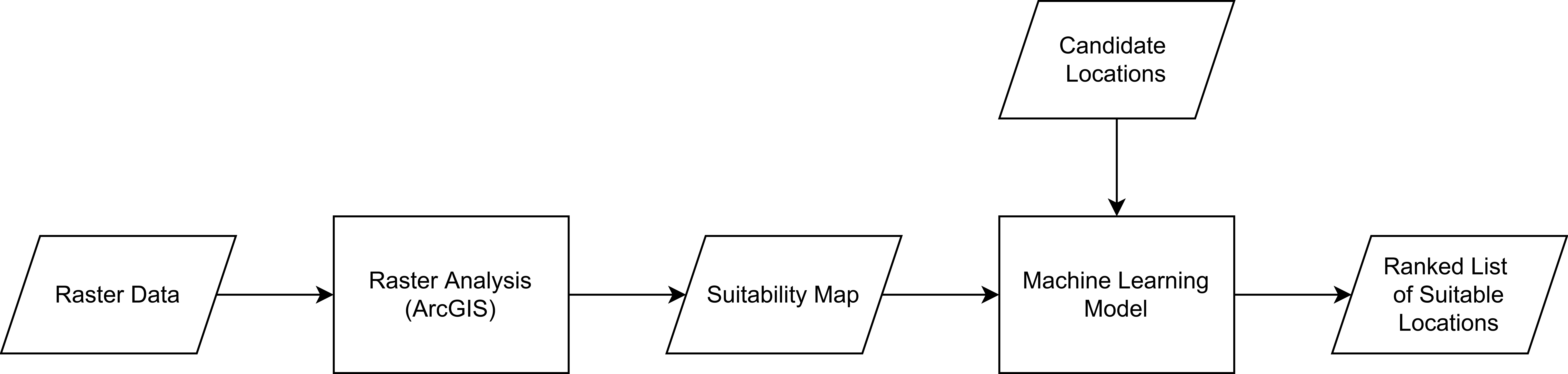}
	\caption{Learning-Based Multi-Criteria Decision Model (LB-MCDM)}\label{fig_model}
\end{figure}

For industrial facility location problem,\citep{adhikari23} applied a multi-criteria comparative analysis method to assess optimal logistics center locations, considering factors such as transportation access, market availability, and economic zones. This adaptable methodology ensures comprehensive site evaluations tailored to different investors' needs. Tested across 13 villages in the Lubuskie Province, the study demonstrated its effectiveness in improving freight transport efficiency and reducing costs, contributing to regional economic growth. Similarly, \citep{wang23} employed a GIS-based MCDM approach to optimize the bioethanol supply chain, strategically selecting biorefinery locations by utilizing local agricultural residues.  

Fuzzy multi-criteria decision-making (fuzzy-MCDM) extends traditional MCDM by incorporating fuzzy logic, which allows for better handling of uncertainty and vagueness. Unlike conventional methods that rely on precise numerical inputs, fuzzy-MCDM uses fuzzy sets to represent imprecise or subjective information, leading to more flexible and realistic decision-making. For instance, \citep{gazi24} applied a fuzzy-MCDM model to determine optimal hospital locations, while \citep{khoshamooz24} used a fuzzy model for selecting fire station sites in Tehran.  

The AHP method \citep{saaty87} remains one of the most widely used MCDM techniques, employing pairwise comparisons to assign weights and rank criteria based on their relative importance. In forest planning, AHP has been a dominant decision-support tool for the past two decades \citep{xing24}. \citep{tocci22} implemented AHP to identify the most suitable extraction system for oak coppice forests in Italy, demonstrating its effectiveness in forestry decision-making.  

Some studies have combined fuzzy logic with AHP to enhance decision-making. For example, \citep{zabaleta24} utilized fuzzy-AHP to assess 12 criteria for controlled landfill site selection, identifying 16 highly suitable landfill locations. This hybrid approach improves the accuracy of site evaluations by integrating both quantitative data and expert judgment, making it a valuable tool for complex decision-making processes.
Although AHP is commonly used in MCDM systems, it introduces bias by relying on subjective opinions to assign weights to different factors. In contrast, machine learning models mitigate this bias by leveraging their inherent structure and feature selection capabilities. These models can uncover hidden patterns within data, enabling more objective and accurate predictions for optimal site selection. For instance, an ML model was applied to hospital site selection in Malacca Province, Malaysia \citep{almansi22}, yielding promising results and demonstrating its effectiveness in reducing bias and enhancing decision-making. Despite the merits of ML models, there are few literatures in applying ML models to sawmill site selection problem. This research tries to fill the research gap by proposing an ML-based approach to solve the sawmill site selection problem.

\section{Problem Statement}
 A decision maker wants to select a subset of promising sawmill locations denoted by $\mathcal{V}^s$ from the set of $N$ candidate locations $\mathcal{V}^c=\{v_1,v_2,\dots,v_N\}$ based on the location suitability scores such that $\mathcal{V}^s\subset \mathcal{V}^c$. 
The decision maker evaluates the candidate locations using multiple criteria, each represented by a raster layer. A raster layer $\textbf{R}$ is a two-dimensional matrix with $m$ rows and $n$ columns, where each cell $r_{ij}$ contains a value of the spatial feature at location $(i,j)$. Given multiple raster layers $(\textbf{R}_1, \textbf{R}_2, \dots, \textbf{R}_K)$ and corresponding weights $(w_1, w_2, \dots, w_K)$, the weighted sum layer $\hat{\textbf{R}}$ is computed as:
\begin{align}
   \hat{\textbf{R}} &= \sum_{i=1}^{K} w_i \textbf{R}_i 
\end{align}
The weighted sum layer (suitability map) provides location-specific composite scores, which are used to train a machine learning (ML) classifier to predict the suitability of candidate locations in $\mathcal{V}^c$. Since this is a binary classification problem and multiple potential ML classifiers exist, let $\vec{y}_j=(y_1, y_2,...,y_N)\in \{0,1\}^N$ denote the dependent variable for the ML classifier $j \in \mathcal{J}$. Each classifier $j$ predicts $\vec{y}_j$ based on the candidate location input feature matrix $\mathbf{x}$ derived from $\hat{\textbf{R}}$ such that $\vec{y}_j=f_j(\mathbf{x})$ and yields $\vec{\alpha}=(\hat{\alpha}_1, \hat{\alpha}_2, ..., \hat{\alpha}_N)\in [0,1]^N$, the resulting vector of the likelihoods of candidate locations in $\mathcal{V}^c$ being suitable. Then, the best-performing ML classifier out of $J$ classifiers is given by $\vec{y}^{*}=f^{*}(\mathbf{x})$ with likelihoods scores $\vec{\alpha}^*=(\hat{\alpha}_1^{*}, \hat{\alpha}_2^{*}, ..., \hat{\alpha}_N^{*})\in [0,1]^N$, which are used to rank-order candidate locations. The next section outlines the methodology employed to address this problem.

\section{Methodology}
We start by describing our raster dataset, followed by presenting the proposed learning-based decision model.

\begin{figure}
    \centering
    \begin{minipage}{0.3\textwidth}
        \centering
        \includegraphics[width=\textwidth]{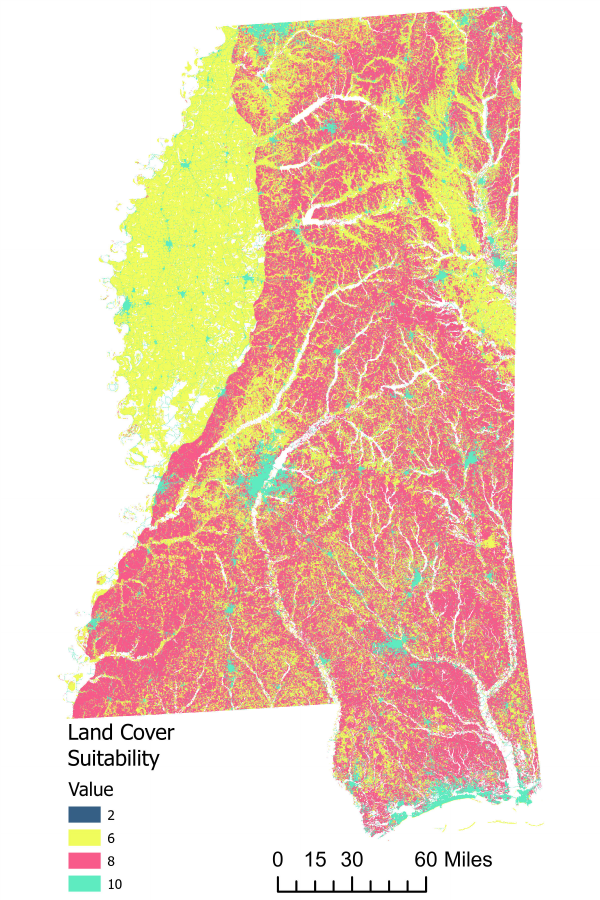}
        \\(a) Land Cover Data
    \end{minipage}
    \hfill
    \begin{minipage}{0.3\textwidth}
        \centering
        \includegraphics[width=\textwidth]{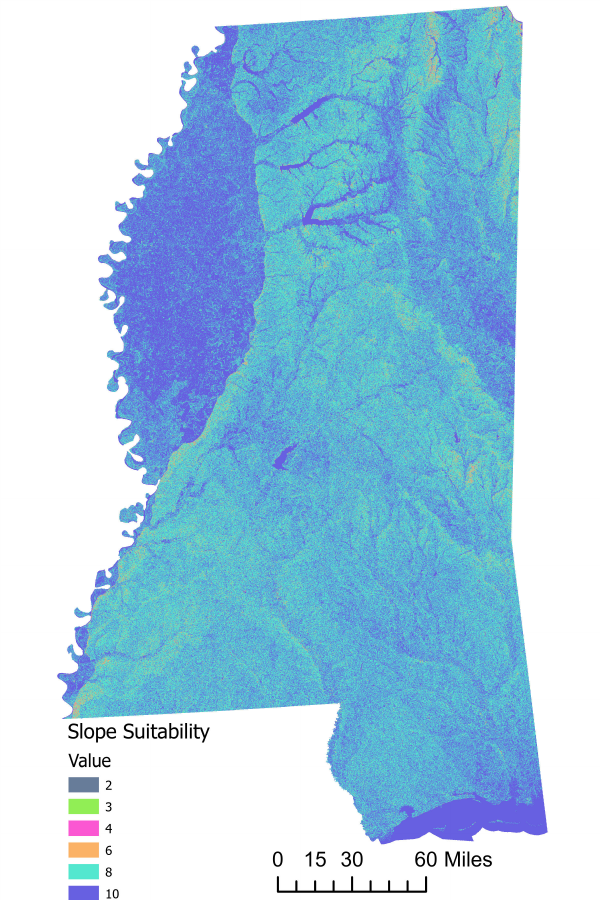}
        \\(b) Slope Data
    \end{minipage}
    \hfill
    \begin{minipage}{0.3\textwidth}
        \centering
        \includegraphics[width=\textwidth]{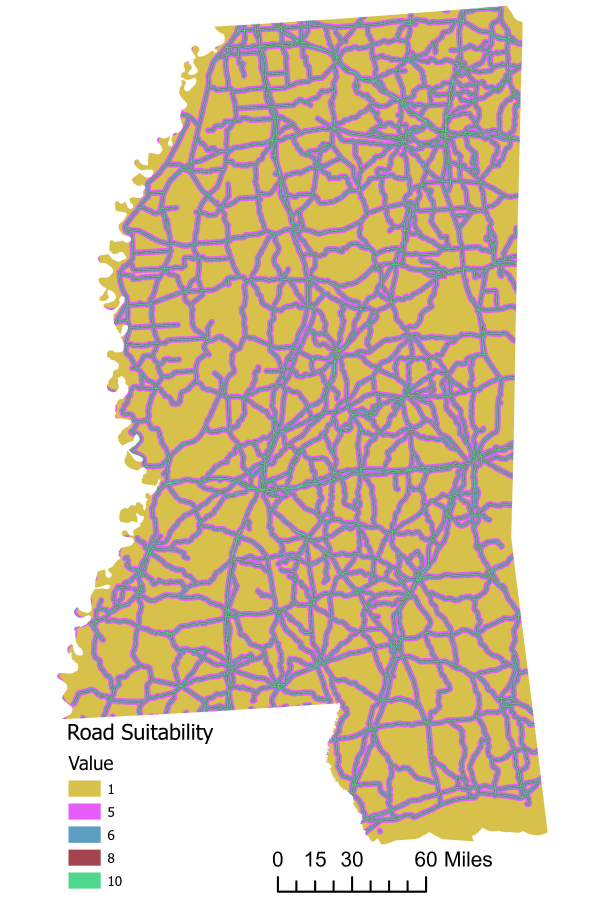}
        \\(c) Road Data
    \end{minipage}
    
    \vspace{5pt}  

    \begin{minipage}{0.3\textwidth}
        \centering
        \includegraphics[width=\textwidth]{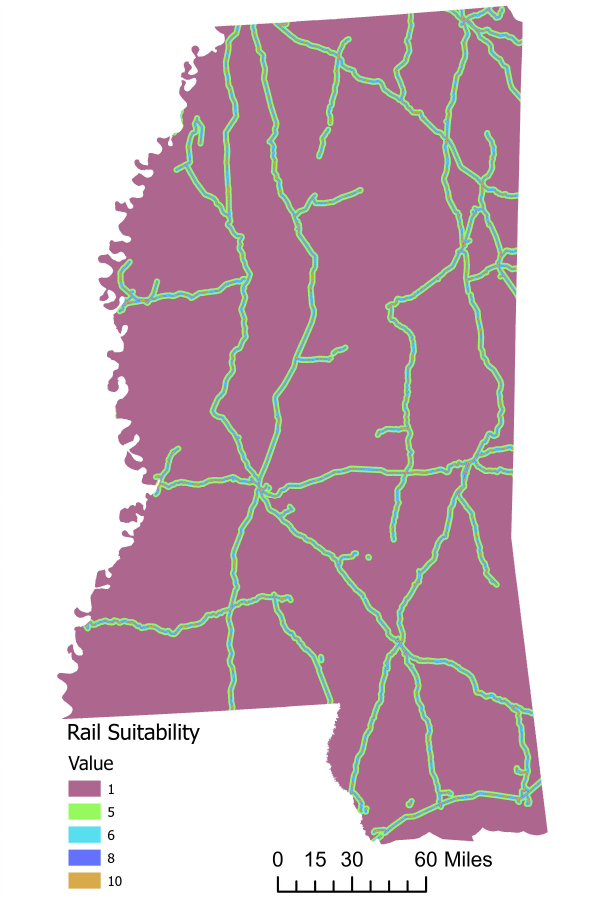}
        \\(d) Rail Data
    \end{minipage}
    \hfill
    \begin{minipage}{0.3\textwidth}
        \centering
        \includegraphics[width=\textwidth]{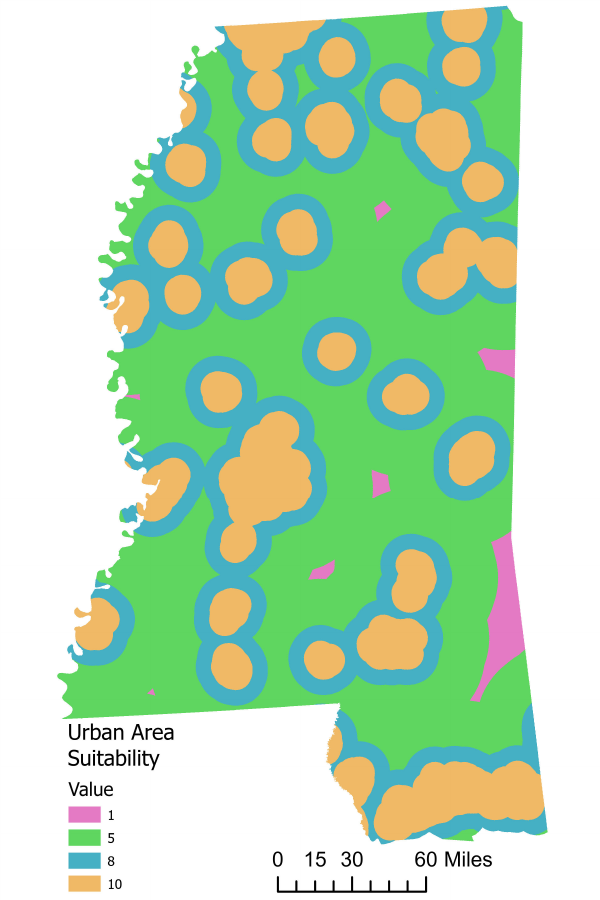}
        \\(e) Urban Data
    \end{minipage}
    \begin{minipage}{0.3\textwidth}
        \centering
        \includegraphics[width=\textwidth]{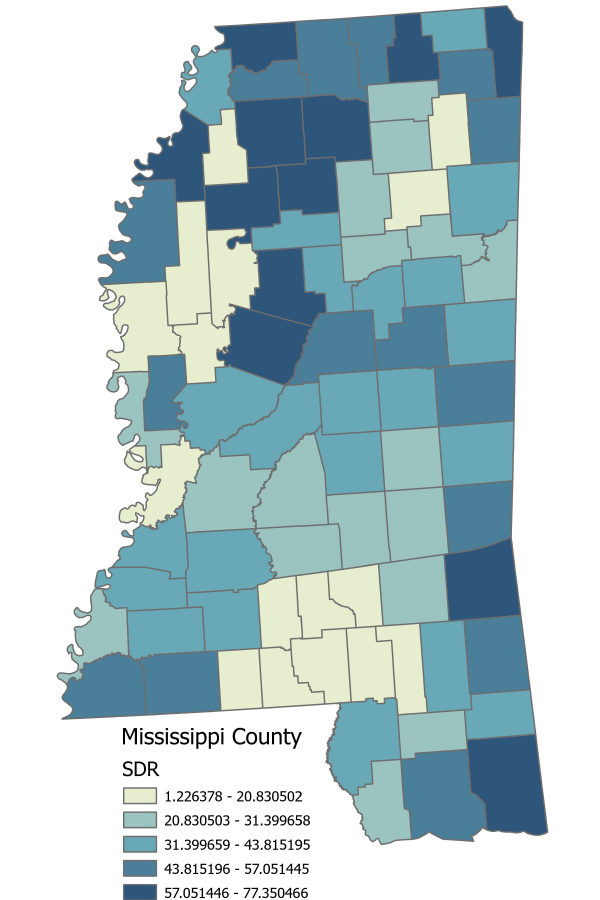}
        \\(f) Supply Demand Ratio
    \end{minipage}
    \caption{(a)-(e) Raster Data }
    \label{Feature_Suitability}
\end{figure}

\begin{figure}
    \centering
    \includegraphics[width=0.75\linewidth]{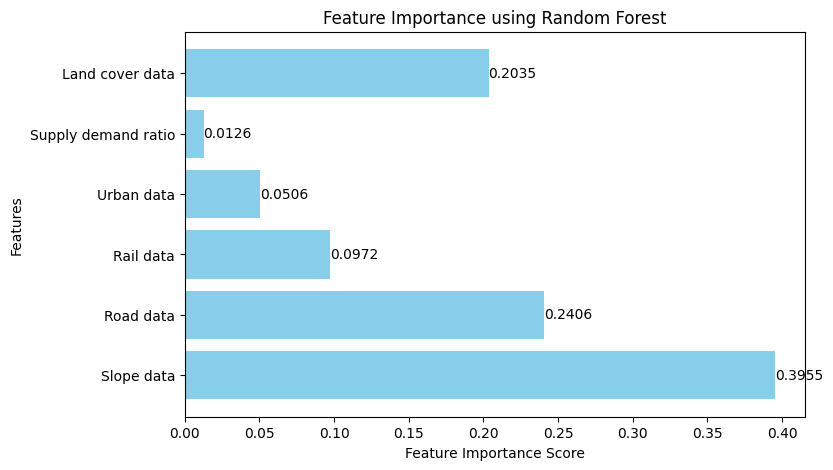}
    \caption{Feature Importance Scores}
    \label{fig:FeatImp}
\end{figure}

\subsection{Raster Data}
The criteria were selected based on prior studies and the relevance of sawmill location, considering geographic and economic factors. Key criteria for site selection include the following raster data:
\begin{enumerate}
    \item \textbf{Land cover data} from the U.S. Census Bureau to exclude unsuitable areas like forests and water bodies, ensuring focus on open and developed lands.
    \item \textbf{Road data} from USGS prioritizes locations within close proximity of roads for cost-effective logistics.
    \item \textbf{Rail data} from USGS prioritizes locations within close proximity of rails for cost-effective logistics.
    \item \textbf{Slope data} from USGS supports slope analysis to favor flat areas, minimizing construction and transportation costs.
    \item \textbf{Urban data} indicates locations in close proximity to urban areas with access to utilities, and populations over one million that offer market access and labor supply. Labor availability is assessed using LAUS data, and Forisk sawmill data guides predictive modeling based on industry trends and existing locations.
    \item \textbf{Supply demand ratio} is a county-level composite construct, indicating the percentage of timber supply consumed by existing sawmills, calculated using the Forisk Sawmill Capacity Database and the Timber Availability Data provided by the MS Department of Agriculture and Commerce.
\end{enumerate}  

\subsection{Learning-Based Multi-Criteria Decision Model (LB-MCDM)} 
Figure \ref{fig_model} depicts the proposed Learning-Based Multi-Criteria Decision Model (LB-MCDM), comprising two stages. 

In Stage 1, raster data is processed using ArcGIS's Raster Analysis feature to create a suitability map, applying the weighting mechanism detailed in Equation 1. The weighted sum layer (suitability map) is sensitive to feature weights. Initially, equal weights are assigned to all spatial features. In Stage 2, the suitability map from Stage 1 is used to generate a large number of random data points with target labels (1 for suitable, 0 for unsuitable), which are then used to train multiple ML classifiers. The best-performing ML classifier is selected based on prediction accuracy, F1 scores, precision, and recall values. During training, the feature importance scores from the best-performing ML classifier are used to adjust the weights. The Raster Analysis is then returned with these updated weights, producing a revised suitability map with a new feature matrix and possibly different target labels. The ML classifier is retrained to generate the final ranked list of candidate locations.

\section{Preliminary Results}

Evaluating the suitability of various features is a critical aspect of site selection. Figure~\ref{Feature_Suitability} presents a comprehensive visualization of each feature on a suitability scale of 10. In terms of land cover, developed open space and forested areas are considered highly suitable, whereas water bodies and wetlands are deemed unsuitable. Terrain suitability is assessed based on slope, with flatter areas being more favorable than steep regions. Proximity to roads and rail networks significantly enhances suitability, as accessibility is crucial for transportation and logistics. Additionally, the figure highlights suitability based on distance from urban areas, underscoring the advantages of proximity to labor, utilities, and markets.

\begin{figure}
    \centering
    \includegraphics[width=0.40\linewidth]{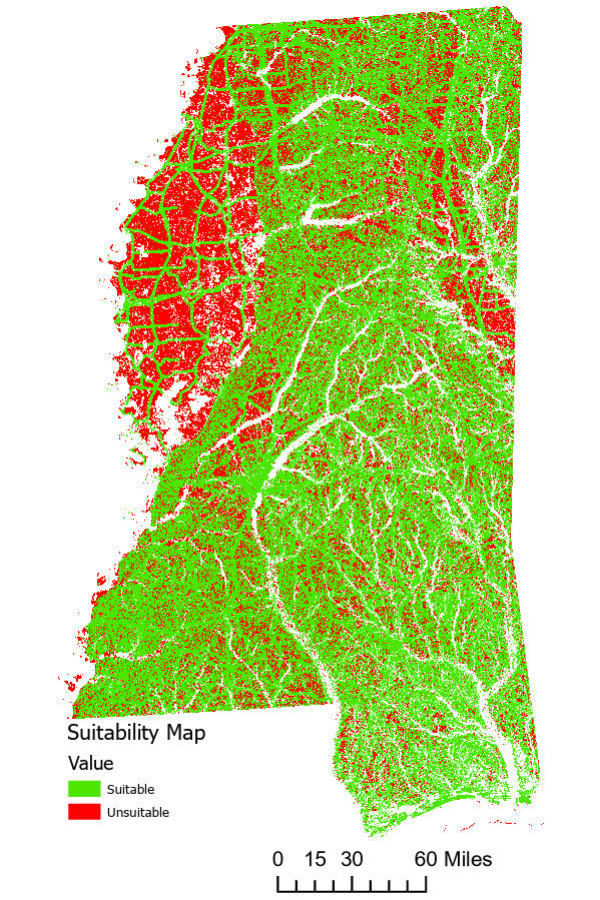}
    \caption{Suitability Map}
    \label{fig:SuitMap}
\end{figure}

Calculating feature importance is essential, as the most influential factors significantly impact the outcomes of the suitability map. The Random Forest classifier assigns importance scores to all features, as shown in Figure~\ref{fig:FeatImp}. According to the results, slope is the most critical site selection criterion, with an importance score of 0.3955, followed by road data (0.2406) and land cover data (0.2035). In contrast, the supply-demand ratio is the least influential factor, with an importance score of 0.0126. The suitability map generated using feature importance scores is presented in Figure~\ref{fig:SuitMap}. 

\section{Conclusion}
Site selection plays a crucial role in establishing a sawmill, yet there has been no dedicated research on selecting optimal sawmill locations. While various site selection studies exist, most rely on the analytic hierarchy process (AHP), which has limitations in accurately determining feature weights due to its reliance on expert judgment. This study introduces a data-driven approach that integrates machine learning (ML) techniques for weight calculation, offering a more objective, scalable, and reliable alternative.  

The goal of this research was to develop an ML-integrated geospatial analysis framework for sawmill site selection. The results demonstrated strong reliability in identifying suitable locations across Mississippi. By leveraging feature importance scores, this study provides a structured decision-making framework that enhances accuracy and efficiency in site selection.  

Future research could explore different ML models, incorporate larger datasets, and expand the study to broader geographical regions, further improving the robustness and applicability of this approach.

\section*{Acknowledgements}
Funding for this research was made possible by the U.S. Department of Agriculture’s (USDA) Agricultural Marketing Service through grant 23FSMIPMS1012. Its contents are solely the responsibility of the authors and do not necessarily represent the official views of the USDA.

\end{document}